\documentclass[10pt,twocolumn,letterpaper]{article}

\usepackage{iccv}
\usepackage{times}
\usepackage{epsfig}
\usepackage{graphicx}
\usepackage{amsmath}
\usepackage{amssymb}
\usepackage{multirow}
\usepackage{booktabs}
\usepackage{arydshln}

\usepackage[accsupp]{axessibility} %
\usepackage{dsfont}
\usepackage{pifont}%

\newcommand{\xmark}{\ding{55}}

\usepackage{color}
\definecolor{rowblue}{RGB}{220,230,240}
\definecolor{myorchid}{RGB}{150,10,30}
\definecolor{myblue}{RGB}{10,30,250}
\definecolor{mygreen}{RGB}{10,120,10}
\definecolor{mypurple}{RGB}{120,0,120}
\definecolor{mymaroon}{RGB}{128,0,0}
\definecolor{myorange}{RGB}{255,69,0}
\definecolor{myteal}{RGB}{20,150,200}

\newcommand{\myparagraph}[1]{\vspace{0pt} \noindent \textbf{#1}}
\newcommand{\mysubsection}[1]{\vspace{1mm}\noindent{\it #1}}

\usepackage[pagebackref=true,breaklinks=true,letterpaper=true,colorlinks,bookmarks=false]{hyperref}

\iccvfinalcopy %

\def\iccvPaperID{25} %

\ificcvfinal\fi

\begin{document}

\title{Contrastive Feature Loss for Image Prediction}

\author{
Alex Andonian$^{1,3}$ \hspace*{0.5em}
Taesung Park$^{2,3}$ \hspace*{0.5em}
Bryan Russell$^{3}$ \hspace*{0.5em}
Phillip Isola$^{1}$ \hspace*{0.5em}
Jun-Yan Zhu$^{3,4}$ \hspace*{0.5em}
Richard Zhang$^{3}$ \\
$^1$MIT \hspace*{1em} $^2$UC Berkeley \hspace*{1em} $^3$Adobe Research \hspace*{1em} $^4$CMU
}

\maketitle
\ificcvfinal\thispagestyle{empty}\fi

\begin{abstract}

Training supervised image synthesis models requires a critic to compare two images: the ground truth to the result. Yet, this basic functionality remains an open problem. A popular line of approaches uses the L1 (mean absolute error) loss, either in the pixel or the feature space of pretrained deep networks. However, we observe that these losses tend to produce overly blurry and grey images, and other techniques such as GANs need to be employed to fight these artifacts. In this work, we introduce an information theory based approach to measuring similarity between two images. We argue that a good reconstruction should have high mutual information with the ground truth. This view enables learning a lightweight critic to ``calibrate'' a feature space in a contrastive manner, such that reconstructions of corresponding spatial patches are brought together, while other patches are repulsed. We show that our formulation immediately boosts the perceptual realism of output images when used as a drop-in replacement for the L1 loss, with or without an additional GAN loss.
Code is available at {\footnotesize\url{https://github.com/alexandonian/contrastive-feature-loss}.}
\vspace{-4mm}
\end{abstract}

\section{Introduction}

\begin{figure}[t]
    \centering
    \includegraphics[width=1.0\linewidth]{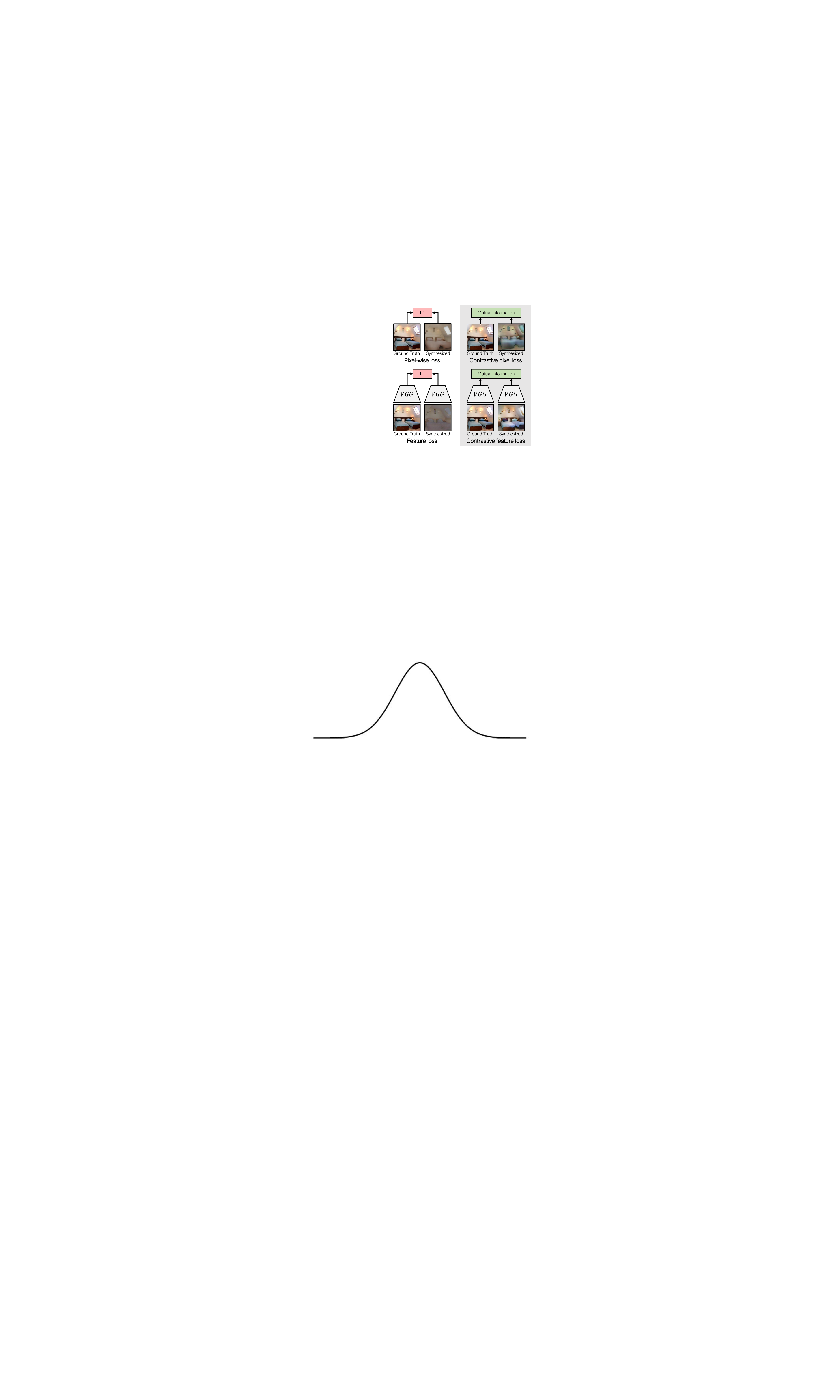}
    \caption{{\bf Patchwise contrastive perceptual loss.} 
    Existing losses for comparing synthesized to ground truth patches use a fixed distance metric such as L1, either in the pixel space~\cite{isola2017image} or some feature space~\cite{chen2017imagesynth,wang2018pix2pixHD,park2019SPADE}. Our method aims to replace the L1 or the L2 regression loss with a loss based on mutual information. Namely, our method learns to group the predicted patch and the ground truth together, distinct from other patches. Our method can be used in either the pixel space or the feature space to improve visual realism. 
    }
    \label{fig:teaser}
    \vspace{-10pt}
\end{figure}

A fundamental requirement for image prediction tasks is an effective loss function, to judge whether the synthesized results are ``close'' to the ground truth.
Straightforward measures -- such as the Hamming distance or Euclidean L2 distance -- do not work well with high-dimensional and structured signals, such as images.

A predominant approach addressing the high-dimensionality of images is to take advantage of the emergent perceptual similarity found in deep network activations, commonly known as the ``perceptual loss'' or feature matching loss~\cite{dosovitskiy2016inverting,ulyanov2016texture,gatys2016image,johnson2016perceptual}.
As seen in Figure~\ref{fig:teaser} (bottom-left), this loss is obtained by computing the L1 distance between the internal activations of a pretrained network to judge the similarity of two images.
Even though the network is pretrained on a seemingly unrelated task (ImageNet~\cite{deng2009imagenet} classification), the corresponding feature space has been shown to be ``unreasonably'' effective in matching human perception~\cite{zhang2018unreasonable} even compared to a rich line of hand-engineered similarity metrics~\cite{wang2004image,zhang2011fsim,mantiuk2011hdr}.

However, despite the success of deep features used for perceptual metrics and
losses~\cite{wang2018pix2pixHD,park2019SPADE,abdal2019image2stylegan,shocher2020semantic} in supervised image synthesis tasks, the original question of how to measure similarity between images has not been fully resolved. In other words, the representations of the signals have been transformed, but an L1 or L2 loss is ultimately still used to judge similarity.

We observe that using the L1 distance metric for image synthesis tasks often results in undesirable visuals that are too smooth or desaturated, even when used in deep feature space (Figure~\ref{fig:teaser} bottom-left). We hypothesize that this is because the L1 loss estimates the median of the multi-modal distribution of the ground truth images. For example, the bed in Figure~\ref{fig:teaser} could be in various styles, and L1 finds the median of all, which turns out to be flat and desaturated. 

In this work, we view structural similarity from the perspective of mutual information.
We posit that a good synthesis is one that is easily associated to the ground truth instance, even if they do not match pixel-by-pixel or feature-by-feature.
While directly estimating mutual information is intractable, recent techniques in unsupervised representation learning~\cite{he2019moco,chen2020simple,oord2018infonce} can be adapted for this purpose.
Namely, we use contrastive learning, which learns an embedding to associate corresponding signals, in \textit{contrast} to other samples in the dataset.
In our case, corresponding signals are patches at the same spatial location in ground truth and synthesized images, and other samples are patches at different spatial locations in the dataset. Furthermore, we find that defining the contrastive loss bidirectionally -- ground truth as the target in one direction, and the synthesized as the target in the other -- improves performance.

We investigate whether this contrastive loss formulation serves as a suitable ``drop-in'' replacement for the standard L1 loss when applied directly on both patches of raw pixels and features extracted by common CNN backbones (Figure~\ref{fig:teaser} right). There are several reasons why a contrastive formulation provides an attractive alternative to L1 loss. In particular, since contrastive learning aims to maximize mutual information, it may potentially capture more desired structure compared to L1. Moreover, the most prominent approaches to contrastive loss lend themselves to being made adaptable to particular data domains, allowing for ``calibration'' on top of a pretrained feature space via a critic function. That is, a small 1-2 layer network is trained on top to select for features, with the practical purpose of emphasizing those that are important for the task, and the theoretical purpose of serving as a better estimator for mutual information~\cite{oord2018infonce}.

We note a few key differences between our method and GAN discriminators~\cite{goodfellow2014generative}, which also learn a critic. Whereas discriminators are trained adversarially, ours is trained in a cooperative fashion, with both generator and critic aiming to maximize mutual information. Ultimately, the methods can be considered complementary. The goal of our method is assessing the \textit{similarity between two images}, while GANs evaluate the \textit{realism of a single image}. We show that our contrastive feature loss can be used in conjunction with a GAN loss, performing competitively against SPADE~\cite{park2019SPADE}, which uses a discriminator and the standard perceptual loss.

We test our method across several image translation tasks, purposely choosing tasks with output domains of varying appearances and levels of photorealism: synthesizing photorealistic (Cityscapes~\cite{cordts2016cityscapes} and ADE20K~\cite{zhou2017ADE20K}) and synthetic scenes (GTA~\cite{richter2016gta}) from labels, predicting depth maps from indoor images~\cite{silberman2012nyuv2}.
Our experimental results show that across these image translation tasks, our contrastive loss formulation serves as a strong replacement for the L1 loss as a distance metric, especially when applied to the embedding space of trained deep networks.

\section{Related Work}
Our work is related to prior work on image similarity, image synthesis, and contrastive feature learning. 

\myparagraph{Image similarity.} Similarity metrics aim to measure how ``close'' two images are. The simplest method is taking a point-wise difference, using Euclidean $\ell_2$, Manhattan $\ell_1$, or PSNR. Such methods do not have the capacity to consider joint statistics. As such, methods developed in the image processing community, such as SSIM~\cite{wang2004image}, FSIM~\cite{zhang2011fsim}, and HDR-VDP-2~\cite{mantiuk2011hdr} build models on top of patches, taking into consideration the human visual system. These methods are effective in situations where structural ambiguity is not a factor, and factors such as photometric changes dominate. However, in many image translation problems, such as translating labels into photorealistic images, synthesizing structure is the predominant factor. Recently, the computer vision community has found that taking distances in the internal embedding space of a deep network, referred to as a ``perceptual loss'' or feature matching loss~\cite{johnson2016perceptual,dosovitskiy2016inverting,gatys2016image,ulyanov2016texture}, works remarkably well for image synthesis~\cite{chen2017imagesynth}. Improvements include adding transformations to allow for more spatial flexibility~\cite{rott2018deformation,mechrez2018contextual}, calibrating the embeddings using human perceptual judgments~\cite{zhang2018unreasonable}, or architectural changes~\cite{ding2020image,gu2020pipal,zhang2019making}. Nonetheless, the backbone remains a network that was trained for a completely different task and data distribution, typically ImageNet classification~\cite{deng2009imagenet}. In our work, we investigate whether it is possible to move beyond a frozen backbone, and \textit{adapt} the feature matching loss to the data and task distribution at hand.

An aspect of image synthesis algorithms is also how real the image looks, in isolation. A natural method of enforcing realism is with a learned discriminator, or GAN loss~\cite{goodfellow2014generative}. Unlike similarity metrics, which compare two images, a discriminator looks at one image in isolation, in a ``no-reference'' fashion, and evaluates it. In image synthesis, similarity and realism are both factors that should be be accounted for, and are sometimes at odds~\cite{blau2018perceptdist}. Thus, discriminators and similarity metrics are often combined in image translation systems in an additive manner~\cite{isola2017image,zhu2017unpaired,wang2018pix2pixHD,park2019SPADE}. Recent works have also explored the coupling between the two ideas, such as re-using the discriminator as a perceptual loss~\cite{wang2018pix2pixHD,wang2018perceptual,park2019SPADE}, as a learned texture descriptor~\cite{park2020swapping}, combining their architectures together~\cite{sungatullina2018image}, or a training schedule that anneals from one to the other~\cite{sun2020progressively}. We show that our method is complementary to GANs, and can be used effectively in conjunction with a learned discriminator.

\vspace{1mm}

\myparagraph{Contrastive feature learning.} 
Contrastive feature learning has primarily looked at learning strong feature representations without supervision for downstream visual recognition tasks~\cite{chen2020simple,he2019moco,misra2020pirl,oord2018infonce,wang2020hypersphere,wu2018unsupervised}. 
These approaches optimize a loss that aims to map a query feature to a positive target, which is obtained via data augmentation~\cite{chen2020simple,he2019moco,misra2020pirl} or leveraging a cross-modal signal~\cite{miech19endtoend,morgado2020avid,tian2020cmc}, and contrast against a set of negative targets. 
Then, a given learned feature representation can be applied to downstream recognition tasks with minimal supervision. 
Our approach is inspired by these efforts, but our goal is different -- we seek to leverage contrastive feature learning for image prediction. 
Our work is related to CUT~\cite{park2020contrastive}, as both use a contrastive loss. However, while CUT asks the output to correspond to the input, we ask the network to synthesize details close to the ground truth image.
Furthermore, we find that a direct implementation of CUT in the paired data domain suffers from bad local minima due to difficult optimization and show that our proposed \emph{bidirectional} formulation effectively mitigates this problem.
Furthermore, since our feature encoder does not share weights with the generator (as it does in CUT), we are able to leverage pretrained feature encoders to produce further downstream performance gains.
Lastly, while the CUT system would not function without a discriminator, we show our method can operate in both settings, with and without a discriminator.

\section{Methods}
\def\generator{G}
\def\encoder{F}
\def\mlp{H}
\def\ginput{X}
\def\goutput{\hat{Y}}
\def\groundtruth{Y}
\def\goutputpatch{\hat{y}_{\text{p}}}
\def\groundtruthpatch{y_{\text{p}}}
\def\plausible{\mathcal{Y}}
\def\MI{\mathcal{I}}
\def\prob{P}
\def\contrastive{\ell_\text{contrastive}}
\def\dist{s}
\def\patch{v}
\def\patchpositive{v^{+}}
\def\patchnegatives{v^{-}}
\def\temp{\tau}
\def\dataset{\mathcal{D}}
\def\discriminator{D}
\vspace{-2mm}
We consider problems where a generator $G$ learns a mapping from input $X\in \mathds{R}^{H\times W\times C_X}$ to $Y\in \mathds{R}^{H\times W\times C_Y}$. For example, for synthesizing photo from a semantic layout, the input is a label map with resolution $H\times W$ and $C_X$ classes, and the output is an RGB image ($C_Y=3$). The generator can be any function approximator, for example a deep convolutional neural network. Our goal is to measure similarity between a ground truth signal $\groundtruth$ and its reconstruction $\goutput=\generator(\ginput)$. 
In this section, we motivate and introduce our bidirectional PatchNCE loss for comparing the similarity of two signals. Moreover, we describe and compare with existing traditional strategies for measuring signal similarity.

\subsection{Mutual information maximization}
\vspace{-2mm}

Often, due to the inherent ambiguity of the task, for a given input signal $\ginput$, there is not just a single ground truth signal but a set of {\em perceptually plausible} output signals $\plausible$. 
For example, given the task of generating an image given an input segmentation mask, a ``car'' mask may correspond to any number of satisfactory distinct instances of the same category. 
As the desired output is often ill-defined from the input signal alone, we seek to define a loss function comparing a generated result $\goutput$ with a ground truth signal $\groundtruth\in\plausible$ that allows for such ambiguity in the signal. 

An example of a poor loss function is per-pixel regression, such as Euclidean $\ell_2$ loss. Using such a loss will average over the outcomes $\mathcal{Y}$, producing a blurry result. This is undesirable, as there is heavy \textit{loss of information}. A blurry result cannot be associated with the ground truth image, relative to other images. We hypothesize that a property of a good reconstruction
is \textit{high mutual information} $\MI$ with the ground truth,
\begin{equation}
    \MI{\left(\goutput; \plausible\right)} = \sum_{\groundtruth\in\plausible} \prob{(\goutput,\groundtruth)} \log{\left(\frac{\prob{(\goutput,\groundtruth)}}{\prob{(\goutput)}\prob{(\groundtruth)}}\right)},
\end{equation}
where $\prob$ denotes probability distributions over random variables. 
Note that if the set of plausible signals $\plausible$ is a single instance, and reconstruction $\goutput$ perfectly matches this, then the mutual information is infinite.

In many cases, we are given a dataset $\dataset = \{(\ginput,\groundtruth)\}$ of input and output ground truth pairs, which contains a \textit{single} observation of the plausible results. We seek a generator $G^{*}$ that maximizes the mutual information of reconstructions and ground truth images, within the dataset, using the following objective,
\begin{equation}
\generator^{*} = \arg\max_{\generator} \; \mathds{E}_{(X,Y)\in\dataset} \; \MI{\left(\generator(\ginput); \groundtruth)\right)}
\end{equation}

\noindent Computing the mutual information directly is intractable, as it requires enumerating all possible outcomes and calculating the joint probability. We use recent advances in the unsupervised learning literature, described below~\cite{he2019moco}.

\begin{figure}[t]
    \centering
    \includegraphics[width=1.0\linewidth]{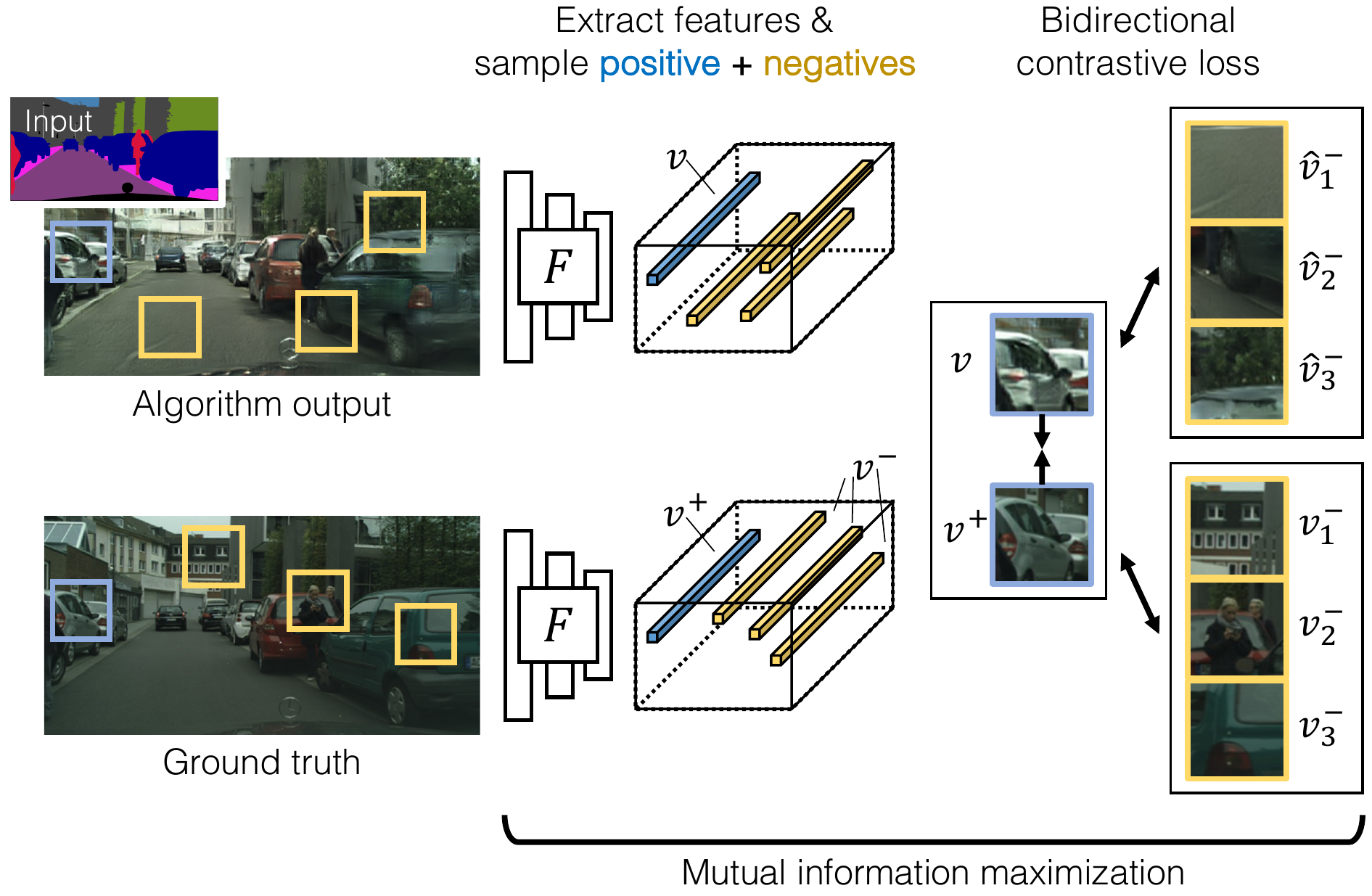}
    \caption{{\bf Bidirectional patchwise contrastive perceptual loss.} 
    Our approach compares spatial patches in a generated signal to patches in a corresponding ground truth. 
    For the depicted label$\rightarrow$image generation task, previous work uses L1 or L2 loss on a feature representation ($\encoder$ can be a frozen feature extractor or identity function to represent raw pixels).
    Based on mutual information, our loss encourages the encoding of corresponding patches in the generated and ground truth signals to be close (blue patches) and non-corresponding patches (yellow) to be far.
}
    \label{fig:approach}
    \vspace{-5mm}
\end{figure}

\subsection{Bidirectional patchwise contrastive loss}
\vspace{-1mm}
We use a noise contrastive estimation framework ~\cite{gutmann2010noise} as a means to maximizing mutual information between reconstruction and ground truth. The idea is that if a reconstruction looks ``similar'' to the ground truth, the two should be able to be embedded closely, relative to other images in the dataset.
Furthermore, we note that for an \textit{image} to be reconstructed well, each \textit{patch} within the image should look similar to the corresponding patch in the ground truth. Thus, we can operate on a patchwise level, which enables denser supervision for generator $\generator$.

We realize this intuition with a \textit{patchwise, contrastive loss}.
Let $\goutputpatch, \groundtruthpatch \in \mathds{R}^{H_\text{patch}\times W_\text{patch} \times C_Y}$ be patches from the two signals, in the same spatial location.
Let $\patch = \encoder(\goutputpatch), \patchpositive = \encoder(\groundtruthpatch) \in \mathds{R}^{D}$ be the $D$-dimensional embedding vectors of spatially-corresponding generated and ground truth patches, and $\patchnegatives \in \mathds{R}^{N\times D}$ be the embedding on the $N$ other patches (``negatives'') from the dataset. In this work we investigate two classes of encoder $\encoder$: (1) a simple linear projection of a patch of pixels, and (2) a pretrained frozen deep network. The loss is defined as an (N+1)-way classification problem, with logits proportional to the similarity between embedded patches,
\vspace{-1mm}
\begin{equation}
\begin{split}
    & \contrastive(\patch,\patch^{+},\patch^{-}) = \\
    & -\log\frac{\exp{(\dist(\patch,\patch^{+})/\temp)}}{\exp{(\dist(\patch,\patch^{+})/\temp)} + \sum_{n=1}^N  \exp{(\dist(\patch,\patch^{-}_n)/\temp)} },
\label{eqn:contrastive}
\end{split}
\end{equation}
where $\dist(\patch_1, \patch_2) = \patch_1^\text{T} \patch_2$ returns the similarity between two encoded patch signals as a dot product, and $\temp$ is a parameter to control temperature. This objective realizes our intuition. If a patch is exactly reconstructed, then the embeddings will have perfect similarity and the numerator will be maximized. If the patch is approximately reconstructed, it obtains a high score if it can be easily associated to the ground truth, relative to the non-matching ``negative'' patches.

\vspace{1mm}
\myparagraph{Multiscale/Multilayer implementation.}
In the case where $\encoder$ is a simple linear layer, patches of pixels are cropped at multiple scales and passed through the layer.
In contrast, when using a deep network as our encoder, we do not literally crop out patches and encode.
Rather, for computational efficiency, we pass the full images $\goutput$ and $\groundtruth$ into a network, producing a stack of features. The feature activation at a given spatial location and layer corresponds to a feature representation of a patch in the input image. We illustrate this setup in Figure~\ref{fig:approach}. The size of the patch depends on the receptive field, depending on the architecture, and the layer of the network. Thus, by taking activations at multiple layers of the network, we are able to compute the patchwise contrastive loss at different spatial scales.

More concretely, images $\goutput, \groundtruth$ are extracted into sets of $L$ feature tensors $\{\hat{V}_l, V_l\}_{l=1}^{L}$, where $L$ is a pre-specified number of layers. Each feature tensor is the output of the $l^\text{th}$ layer of encoder $\encoder$, and a small projection head (e.g. a linear layer or two-layer MLP) $\mlp$. There is a slight abuse of notation; the composition of $\mlp$ and $\encoder$ corresponds to $\encoder$ in the previous subsection. Adding a small MLP was shown by Chen et al.~\cite{chen2020simple} to improve performance, so we found this practice compatible with our framework. The shape of tensor $V_l \in \mathds{R}^{S_l \times D_l}$ is determined by the architecture of the network, where $S_l$ is the number of spatial locations of the tensor.
We index into the tensor with notation $v_l^s \in \mathds{R}^{D_l}$, which is the $D_l$-dimensional feature vector at the $s^\text{th}$ spatial location.
We denote $\bar{v}_l^s \in \mathds{R}^{(S_l - 1) \times D_l}$ as the collection of feature vectors at all other spatial locations.

Our loss on a pair of images is written as follows.
\begin{equation}
    \mathcal{L}_\text{contrastive}(\goutput, \groundtruth) = \sum_{l=1}^{L} \sum_{s=1}^{S_l} \contrastive(\hat{v}_l^s, v_l^s, \bar{v}_l^s)
\end{equation}

\myparagraph{Bidirectional PatchNCE Loss.} Drawing inspiration upon classic work in image similarity~\cite{simakov2008summarizing}, we investigate a symmetric ``bidirectional'' variant of the PatchNCE loss. We pursue this with the intuition that not only should generated patches contrast with non-corresponding in training patches, the corresponding training patch should similarly contrast with non-corresponding \textit{generated} patches. We find that this leads to more stable training behavior, faster convergence, and avoids degenerate solutions. We find this to consistently hold true across different settings. Here, a second contrastive term is added with the roles of the generated and ground truth patches reversed, as shown below:
\vspace{-3.5mm}
\begin{align}
    &\mathcal{L}_\text{contrastive}(\goutput, \groundtruth) =\\
    &\frac{1}{2}\sum_{l=1}^{L} \sum_{s=1}^{S_l} \contrastive(\hat{v}_l^s, v_l^s, \texttt{sg}(\bar{v}_l^s)) + 
    \contrastive({v}_l^s, \hat{v}_l^s, \texttt{sg}(\hat{\bar{v}}_l^s)) \nonumber
\end{align}
where \texttt{sg} function indicates a stop gradient operation (i.e., gradients are prevented from flowing through the negatives). In the representation learning context, SimSiam~\cite{chen2020exploring} uses this technique to help prevent degenerate solutions. All experiments take advantage of this bidirectional variant, unless noted otherwise.

\myparagraph{Final contrastive objective.} Given a dataset $\dataset = \{(\ginput,\groundtruth)\}$ of input and output ground truth pairs, we obtain the generator $G^{*}$ with the following objective,
\begin{equation}
    \generator^{*} = \arg\min_{\generator} \min_{\mlp} \mathds{E}_{(\ginput,\groundtruth)\in\dataset} \; \mathcal{L}_\text{contrastive}(\generator(\ginput), \groundtruth).
    \label{eqn:final}
\end{equation}
As encoder $\mlp$ and generator $\generator$ are neural networks, then the entire system is differentiable and can be optimized jointly using stochastic gradient descent.

\begin{figure*}[t!]
    \centering
    \includegraphics[width=1.0\linewidth]{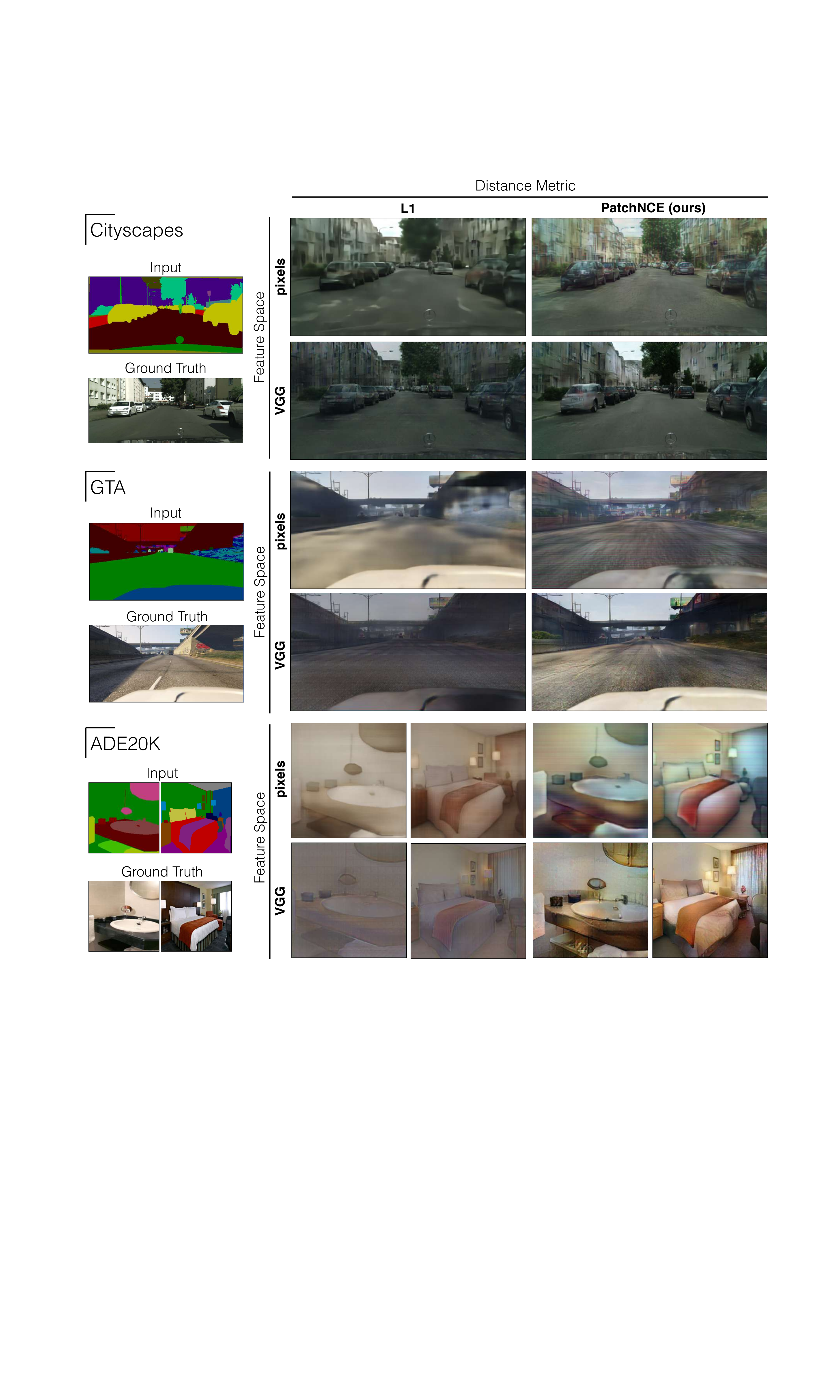}
    \caption{We compare our results against typical regression losses used in supervised images synthesis tasks, such as L1 in pixel space~\cite{pix2pix2017} and the feature space of the pretrained VGG network~\cite{chen2017imagesynth,wang2018perceptual,park2019SPADE}, on the Cityscapes\cite{cordts2016cityscapes}, GTA\cite{richter2017playing}, and ADE20K\cite{zhou2017ADE20K} dataset. Because the L1 loss encourages regressing toward the median of all possible targets, they end up producing overly blurred or desaturated results. In contrast, our PatchNCE loss, used in either the pixel or feature space, significantly boosts sharpness and colorfulness. Note that other techniques such as progressive growing or GAN are not used in this experiment.}
    \label{fig:cityscapes_gta}
    
\end{figure*}

\vspace{-1.mm}
\subsection{Classic feature matching loss}
\vspace{-1mm}
Our loss has similarities with the classic feature matching loss, or ``perceptual loss''. The loss also passes image $\goutput, \groundtruth$ through a pretrained network, which we designate as $\encoder$, to produce a stack of features $\hat{V}, V$.
The perceptual loss uses the $\ell_2$ function to compare feature activations,
\vspace{-2mm}
\begin{equation}
    \mathcal{L}_\text{classic}(\goutput, \groundtruth) = \sum_{l=1}^{L} \sum_{s=1}^{S_l} \ell_2(\hat{v}_l^s, v_l^s).
\end{equation}
\vspace{-2mm}

The loss is used to optimize a generator, in the following,
\vspace{-1mm}
\begin{equation}
    \generator^{*} = \arg\min_{\generator}
    \mathds{E}_{(\ginput,\groundtruth)\in\dataset} \; \mathcal{L}_\text{classic}(\generator(\ginput), \groundtruth).
\end{equation}

Comparing the optimization above with our method in Equation \ref{eqn:final}, we note two differences.
Firstly, rather than only bringing corresponding patches together, ours pushes them apart from non-corresponding patches at the same time. Secondly, we have a shallow, learnable network $H$ on top of the primary feature extractor $F$, that can select for the more important features for a given task.
\vspace{-1mm}
\subsection{GAN loss}
\vspace{-1mm}
We note that our method provides a lightweight mechanism for calibrating to data, reminiscent of a GAN discriminator. A key difference is that our method judges the similarity between reconstruction and ground truth, whereas a GAN loss judges the ``realism'' of just the reconstruction conditional on the input. It does so by 
learning a discriminator $\discriminator$ that judges whether an input output pair of $(\ginput, \generator(\ginput))$ looks like it belongs to a dataset of input-output pairs $(\ginput, \groundtruth)\in \dataset$. 
\vspace{-2mm}
\begin{equation}
    \mathcal{L}_\text{cGAN}{(\ginput, \goutput,\groundtruth)} = \log \discriminator(\ginput,\groundtruth) + \log(1-\discriminator(\ginput,\goutput))
\end{equation}

The generator aims to ``fool'' the discriminator. The solution to the generator is found via min-max optimization. 
In our experiments, we show that our method can be used both by itself, or in conjunction with a discriminator,
\vspace{-2mm}
\begin{equation}
\begin{split}
    \generator^{*} = \arg\min_{\generator} \min_{\encoder, \mlp} \max_{\discriminator} \mathds{E}_{(\ginput,\groundtruth)\in\dataset}
    \big[ & \mathcal{L}_\text{contrastive}(G(\ginput), \groundtruth) \\
    + \hspace{1mm} & \mathcal{L}_\text{cGAN}(\ginput, \generator(\ginput), \groundtruth) \big].
\end{split}
\end{equation}
\vspace{-12mm}

\section{Experiments}
We evaluate our method on several paired image-to-image translation datasets, for synthesizing both photorealistic and non-photorealistic images, and compare against several baselines, followed by model ablations.

\myparagraph{Network architecture and training details.} As our main interest is investigating the patch-based contrastive loss, we keep the same training procedure and architecture established by Pix2PixHD~\cite{wang2018pix2pixHD} and SPADE~\cite{park2019SPADE}, except the loss term. In more detail, for our generator network, we use the ResNet-based SPADE generator, which pairs well with the multiscale PatchGAN discriminator network (when present).
We investigate several different backbone architectures for the feature extraction network in our contrastive loss (see supp. for complete list), focusing specifically on the VGG19~\cite{simonyan2014very} network standard ``perceptual'' loss. Features extracted by these networks are then reduced to low-dimensional embeddings via a simple linear layer projection head and normalized by their L2 norms.

\myparagraph{Baselines.} 
Our main goal is to compare the effect of the data-driven contrastive loss function as an improved drop in replacement for the ubiquitous L1/L2 loss. Therefore, we compare our approach against simple $L1$ pixel distance loss, $L1$ distance of VGG feature responses~\cite{chen2017imagesynth}, in combination with the GAN loss~\cite{park2019SPADE}, while we use the same network architecture for a fair comparison. Therefore, we do not directly compare our method against prior work \cite{chen2017imagesynth,wang2018pix2pixHD} as they used different architectures. For SPADE, we use publicly available code.

\begin{figure*}
    \centering
    \includegraphics[width=1.0\linewidth]{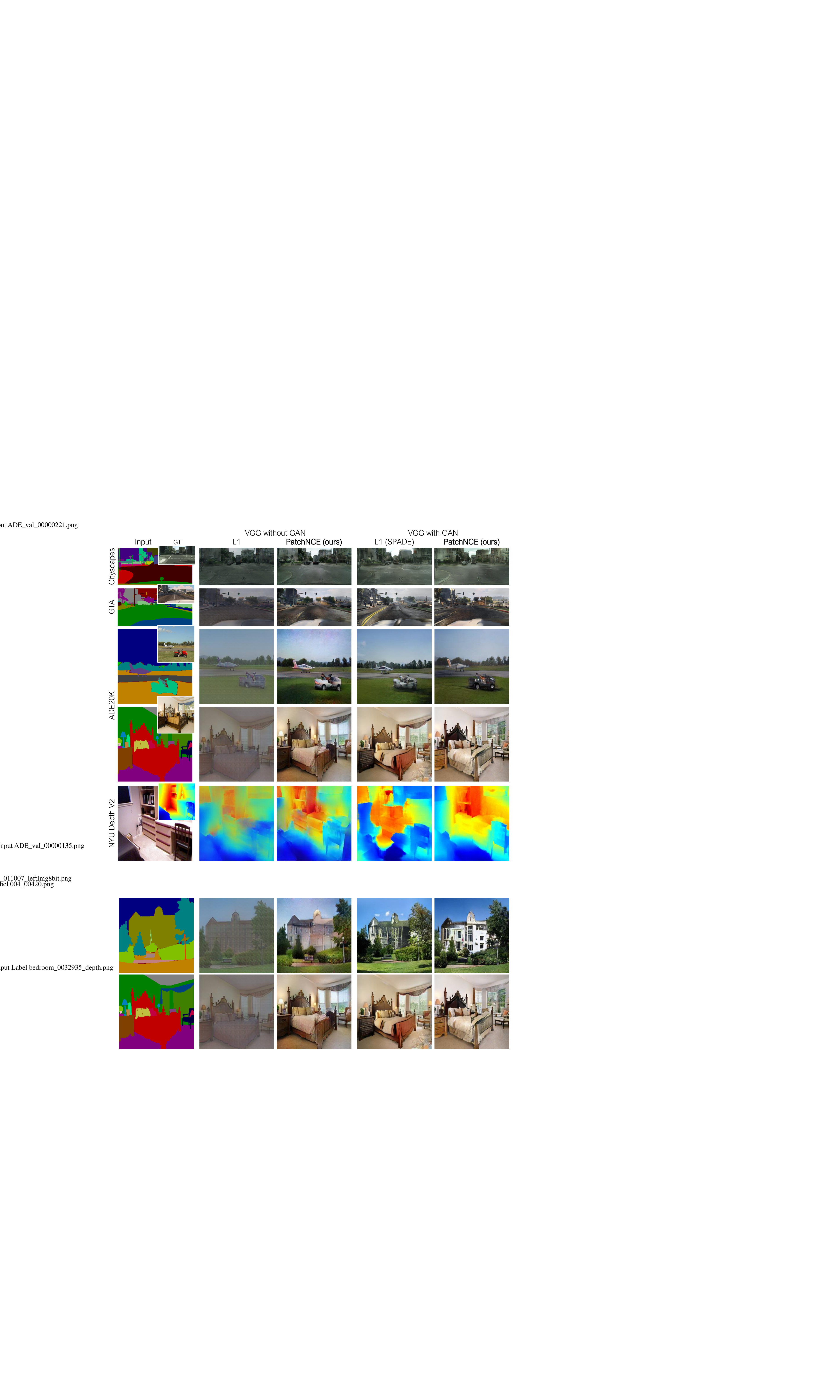}
    \caption{We show the effect of using the PatchNCE loss in comparison to the L1 feature loss on the Cityscapes, GTA, ADE20K, and NYU Depth V2 dataset, without and with the GAN discriminator. Even without GAN, our PatchNCE loss brings out sharpness and saturation, almost matching the quality of SPADE~\cite{park2019SPADE}, which uses GAN as well as the L1 feature matching loss. Adding GAN to PatchNCE further enhances realism by removing small grid artifacts in the outdoor scene (ADE20K, third row) for example.}
    \label{fig:comparison_gan}
    \vspace{-5mm}
\end{figure*}

\subsection{Paired Image-to-Image Translation} 
\myparagraph{Datasets.} We evaluate our method on several datasets with varying amounts of photorealism. 

\mysubsection{Cityscapes}~\cite{cordts2016cityscapes} contains street photographs paired with their corresponding semantic label maps for 2,975 training and 500 validation images. Models learn to translate semantic label maps containing up to a total of 19 unique semantic categories to 512$\times$256 resolution output images.

\mysubsection{GTA dataset}~\cite{richter2017playing} contains synthetic scenes from the widely-popular \emph{Grand Theft Auto V} video game. This dataset is semantically similar to Cityscapes but is larger with 26k training images and 5k testing images. The resolution is same as the Cityscapes dataset.

\mysubsection{NYU Depth V2}~\cite{silberman2012nyuv2} contains video frames of 464 indoor scenes from 3 cities recorded by the Microsoft Kinect RGB-D cameras. Our task is to synthesize depth maps from RGB images at 256x256 resolution. We create an 80\%-20\% split of the 1,449 pre-processed pairs to synthesize colorized depth map from RGB images.

\mysubsection{ADE20K}~\cite{zhou2017ADE20K} consists of challenging indoor and outdoor scenes labeled for 150 different semantic classes. The dataset is split into 20,210 training images and 2,000 validation  images.

\myparagraph{Evaluation protocol.}
We adopt evaluation protocols that are tailored to traditional image translation tasks as well as more domain-specific metrics such as those frequently employed in the pixel labeling and depth estimation literature.
To briefly summarize, we first use the standard Fr\'{e}chet Inception Distance (\textbf{FID}) metric~\cite{heusel2017fid}, which is aimed at assessing the visual quality of generated samples compared to real images by computing the statistics between the distributions of real and generated images estimated by a deep neural network.
In the standard setting, this network is an InceptionV3 pretrained on the ImageNet dataset.
However, FID does not evaluate the correspondence between inputs and outputs. Therefore, for semantic image synthesis tasks of the Cityscapes, GTA and ADE20K datasets, we adopt prior protocol and evaluate segmentation accuracy~\cite{chen2017imagesynth,wang2018pix2pixHD,park2019SPADE}. We report the mean average precision (\textbf{mAP}) and pixel accuracy (\textbf{Acc}).
In order to evaluate the performance of our approach on the NYU V2 depth estimation task, we report the standard absolute relative difference (\textbf{rel}) and RMSE-linear (\textbf{rms}) criteria~\cite{eigen2014depth,karsch2012depth}. 

\vspace{3mm}
\myparagraph{Qualitative evaluation.} As shown in Figure~\ref{fig:cityscapes_gta}, the L1-regression loss in the pixel or the VGG feature space results in desaturated, blurry images, since the information loss in the output image is not penalized. In contrast, our PatchNCE method does not suffer from this same phenomena and produces output images with increased sharpness and color saturation. We observe that the increased sharpness and saturation makes the outputs similar to the results of GAN-based methods like SPADE~\cite{park2019SPADE} (Figure~\ref{fig:comparison_gan}). Moreover it is seen that our PatchNCE can also further benefit from the addition of GAN.
\vspace{-1mm}

\begin{table*}
    \resizebox{1.\linewidth}{!}{
    \centering
    
    \begin{tabular}{cccccccccccccc}
    \toprule
                          \multirow{2}{*}{\shortstack[c]{Feature \\ Space}} & \multirow{2}{*}{\shortstack[c]{Loss}} & \multicolumn{3}{c}{Cityscapes~\cite{cordts2016cityscapes}}                & \multicolumn{3}{c}{GTA~\cite{richter2017playing}}                       & \multicolumn{3}{c}{ADE20K~\cite{zhou2017ADE20K}}                    & \multicolumn{3}{c}{NYU Depth V2~\cite{silberman2012nyuv2}}               \\ \cmidrule(lr){3-5} \cmidrule(lr){6-8} \cmidrule(lr){9-11} \cmidrule(lr){12-14}
                            &           & FID $\downarrow$           & mAP $\uparrow$          & Acc $\uparrow$          & FID $\downarrow$           & mAP  $\uparrow$        & Acc $\uparrow$          & FID $\downarrow$      & mAP $\uparrow$      & Acc $\uparrow$      & FID $\downarrow$           & rel $\downarrow$           & rms $\downarrow$          \\ \midrule
\multirow{2}{*}{Pixel}     & L1                                   & 135.1          & 45.3          & 76.3         & 132.1          & 20.2          & 33.7          & 126.5          & 7.4           & 43.4          & 264.6          & \textbf{0.431} & 1.83          \\
                           &                   PatchNCE                 & \textbf{107.3} & \textbf{48.2} & \textbf{78.2} & \textbf{107.2} & \textbf{23.1} & \textbf{40.1} & \textbf{125.9} &  \textbf{8.7}          & \textbf{46.2}          & \textbf{170.6} & 0.434          & \textbf{1.77} \\ \cdashline{1-14} 
\multirow{2}{*}{VGG}       & L1~\cite{chen2017imagesynth}                   & 76.2           & 60.5          & 81.9          & 62.22          & 31.6          & 47.7          & 74.6           & 20.9          & 63.6          & 103.2          & \textbf{0.410} & \textbf{1.80} \\
                           &                  PatchNCE                 & \textbf{68.3}  & \textbf{64.6} & \textbf{82.4} & \textbf{47.3}  & \textbf{33.6} & \textbf{50.7} & \textbf{48.9}  & \textbf{29.0} & \textbf{69.5} & \textbf{97.2}  & 0.413          & 1.87          \\ \cdashline{1-14}
\multirow{2}{*}{\shortstack{VGG \\ + GAN}} & L1~\cite{park2019SPADE}                   & \textbf{58.7}  & 62.3          & 81.9          & \textbf{36.3}  & 30.8          & 47.3          & \textbf{33.1}  & \textbf{38.5} & \textbf{79.9} & 86.5           & 0.454          & 2.12          \\
                           &                  PatchNCE                 & 66.7           & \textbf{66.1} & \textbf{82.4} & 39.4           & \textbf{34.8} & \textbf{51.9} & 34.9           & 29.1          & 70.8          & \textbf{64.0}  & \textbf{0.444} & \textbf{2.11} \\
                           \bottomrule
                           
\end{tabular}
}
    \caption{Using the PatchNCE loss instead of L1 improves quantitative metrics across different datasets and feature spaces. For each dataset, we compare the results of training with our PatchNCE loss as opposed to the standard L1 regression loss. The experiments are also performed in different feature spaces, including just the pixel space (Pixel), the feature space of a pretrained VGG19 network (VGG), and VGG in combination with a GAN discriminator (VGG + GAN). In absence of GAN (the first four result rows), the advantage of our method over L1 is clear across datasets and various metrics. When GAN is added, we observe that the gap is reduced, likely because the GAN discriminator ``fixes'' the typical artifacts of the L1 regression. Still, our method attains superior correspondence metrics (mAP, Acc, rel, rms) overall. Note that L1 loss in VGG + GAN corresponds to the original SPADE~\cite{park2019SPADE}. Also, the L1 regression loss in VGG space is similar to CRN~\cite{chen2017imagesynth}, except that the SPADE architecture was used for fair comparison. }
    \label{tab:comparison}
    \vspace{-2mm}
\end{table*}
\myparagraph{Quantitative evaluation.} In Table \ref{tab:comparison}, we show quantitative measures comparing our method variants to alternative approaches for two street scene datasets, one natural (Cityscapes) and one synthetic (GTA), a diverse indoor-outdoor dataset (ADE20K), and RGB - colorized depth map dataset (NYU Depth V2). 
Here, we compare our method against baselines using $L1$ pixel regression, $L1$ loss on VGG features as in \cite{chen2017imagesynth}, and also in combination with the GAN loss used in \cite{wang2018pix2pixHD,park2019SPADE}.
We present three variants of our method: the generator trained with only the patch-based contrastive loss applied directly to raw pixels (Pixel PatchNCE), contrastive loss applied to fixed VGG features (VGG PatchNCE), and our formulation that incorporates a GAN discriminator loss (VGG + GAN PatchNCE).
In nearly all cases across these datasets, simply replacing the L1 metric with our contrastive loss formulation brings notable improvements to segmentation metrics (mAP, Acc) and depth estimation metrics (rel, rms). Moreover, our method consistently lowers FID scores for the variants that forgo the GAN discriminator loss, suggesting that the PatchNCE loss helps to bring significant improvements in photo-realism.
For completeness, we note that published CUT~\cite{park2020contrastive} performance on the Cityscapes semantic segmentation metrics (mAP: 24.7, Acc: 68.6) is significantly lower than all other methods in Table \ref{tab:comparison}, underscoring the advantages of using paired data when it is available.

One advantage afforded by our implementation of a contrastive loss over L1 loss is the ability to still take advantage of strong feature extractors pretrained on large datasets while adapting it to new data domains in an efficient way. Specifically, our methods trains a lightweight projection head as it learns to become a better critic (i.e., a better estimator of mutual information). 
Therefore, we hypothesize that this difference should prove to be advantageous when the target domain diverges from natural images.
Indeed, the qualitative and quantitative comparison shown on the GTA and NYU Depth V2 dataset support this hypothesis with our method showing performance improvements on synthesizing CG images or colorized depth maps.

\myparagraph{Importance of Bidirectional PatchNCE and pretraining.}
In Table \ref{tab:ablations} we show the performance of our method under several ablations with and without a GAN discriminator loss. In particular, we compare standard vs bidirectional PatchNCE loss variants and investigate the effect of unfreezing the underlying feature network, or even training it from scratch. Compared to the baseline method where L1 loss is applied to VGG features, all bidirectional variants achieve higher mAP scores than the baseline (60.5), even if the underlying feature extractor is trained from scratch on the target domain. Keeping all other factors fixed, the bidirectional NCE loss performs consistently better than the standard NCE, but the bidirectional NCE loss becomes increasingly important in stabilizing training when the underlying feature extractor is also being trained. The last row in each section of in Table \ref{tab:ablations} most closely resembles a direct application of CUT to paired data with/without a discriminator, and its trailing performance further highlights the advantages of our proposed bidirectional formulation.

\newcommand{\greencheck}{{\color{green}\checkmark}}
\newcommand{\redx}{{\textcolor{red}\xmark}}
\begin{table}[]
    \resizebox{1.\linewidth}{!}{
    \centering
    \small
    \begin{tabular}{cccccc}
    \toprule
    \multicolumn{4}{c}{\textbf{Method}} & \multicolumn{2}{c}{\textbf{Performance}} \\ \cmidrule(lr){1-4} \cmidrule(lr){5-6}
    NCE variant    & GAN   & Pretrained $F$ & Frozen $F$  &FID $\downarrow$ & mAP $\uparrow$ \\
    \hline
    Bidirectional* & \greencheck & \greencheck    & \greencheck & 66.7  & 66.1 \\
    Standard       & \greencheck & \greencheck    & \greencheck & 69.3  & 64.9 \\  
    Bidirectional  & \greencheck & \greencheck    & \redx       & 63.0  & 64.6 \\  
    Standard       & \greencheck & \greencheck    & \redx       & 68.8  & 58.8 \\
    Bidirectional  & \greencheck & \redx          & \redx       & 76.2  & 62.0 \\  
    Standard       & \greencheck & \redx          & \redx       & 87.1 &  57.7 \\
    \cdashline{1-6} 
    Bidirectional* & \redx       & \greencheck    & \greencheck & 68.3  & 64.5 \\
    Bidirectional  & \redx       & \greencheck    & \greencheck & 72.1  & 62.6 \\
    Standard       & \redx       & \greencheck    & \greencheck & 73.8  & 61.6 \\
    Bidirectional  & \redx       & \greencheck    & \redx       & 85.9  & 62.9 \\
    Bidirectional  & \redx       & \redx          & \redx       & 86.0  & 62.7 \\
    Standard       & \redx       & \greencheck    & \redx       & 135.7 & 42.5 \\
    Standard$^\dagger$ & \redx  &  \redx     & \redx       & 143.0 & 39.3 \\
    \bottomrule
    \end{tabular}
    }
    \caption{FID and mAP scores on Cityscapes under different configurations of the VGG-PatchNCE method, including whether the backbone VGG network is frozen and/or pretrained, or if GAN loss is present.
    Bidirectional* forgoes NCE loss applied to the first layer appearing in the original VGG loss implementation (layer 2) and serves as our default configuration. Standard$^\dagger$ is analogous to CUT~\cite{park2020contrastive} without a discriminator adapted for use with paired data.
     }
    \label{tab:ablations}
    \vspace{-5.0mm}
\end{table}

\vspace{-3.0mm}
\section{Conclusion}
\vspace{-3mm}
We have introduced an approach for learning a lightweight image similarity critic via our PatchNCE formulation. 
We have demonstrated that the standard L1 loss frequently applied in pixel or feature space can be replaced by learning a new distance metric based on maximizing information between patches. Our loss formulation achieves competitive or superior results on synthesizing both natural and non-natural images, and works well in raw pixel space or the feature space of standard deep networks, with or without a GAN discriminator. Our approach opens up new avenues for exploring adaptive critics in other signal modalities, such as video and audio.

\myparagraph{Acknowledgments} We thank Yonglong Tian, Dilip Krishnan and Aude Oliva for helpful discussion and feedback. This work was initiated while AA and TP were interns and JYZ was an employee at Adobe. This work was supported in part by an Adobe gift.

{\small
\bibliographystyle{ieee_fullname}
\bibliography{main}
}

\appendix
\section{Additional network architectures and training/evaluation details}

\subsection{Feature spaces}
As described in the main paper, our main experiments focus on two primary feature spaces (1) Pixel space and (2) the VGG19~\cite{simonyan2014very} feature space.

\myparagraph{Pixel Space.} In order to apply our NCE loss to raw pixel space, we begin by cropping the ground truth and generated images into patches of varying sizes: 4x4, 8x8, 16x16 and 32x32 crops with these patch scales separated by strides of 2, 4, 8, 16 pixels respectively. 1024 randomly chosen pixel patches are then ``flattened'' into 1D vectors before passing through a learned linear linear layer to produce the 256 dimensional representation that participates in the NCE loss.

\myparagraph{VGG Feature Space.} Here, we use the pretrained VGG19 network available for download using PyTorch's \texttt{torchvision} utility library containing common pretrained networks. In keeping with the VGG loss present in SPADE~\cite{park2019SPADE}, features are extracted from the output of the ReLUs found at layers 2, 7, 12, 21, and 30. Then, 1024 depth columns/fibres (corresponding to spatial patch locations) are randomly selected from the feature tensors and pass through a simple (learned) linear layer to obtain low dimensional embeddings with size 256.

In both the pixel and VGG cases, these embeddings are L2 normalized to lie on the unit hypersphere before participating in the NCE loss. As with previous works, our NCE loss uses a softmax temperature of 0.07.

\subsection{NCE Patch Selection.}
\myparagraph{Number of negative patches.} We tested our method with the number of negative patches ranging from as few as 256 to as many 4096 and found it fairly robust to this choice. Since the computational cost of more negative patches appeared to outpaced the corresponding improvements in downstream performance, we opted to run experiments with 1024 patches to strike a balance between performance and efficiency.

\myparagraph{Negative mining from same image.} We experimented with taking negative patches from within the same
image vs. negative patches from other images in the same batch and did not observe significant qualitative or quantitative differences. This finding suggests that with enough randomly chosen patches, a sufficient number of proper negatives emerge and allow learning to proceed.

\subsection{Optimization details}
The generator and discriminator architecture of all our methods and baselines follow those of SPADE~\cite{park2019SPADE} for fair comparison. Our models are trained anywhere from 400-1000 epochs using hyperparameter settings similar to \cite{park2019SPADE}, with batch sizes ranging from 12-64 examples, Adam optimizer with learning rate 0.0002, and flip-equivariance data augmentation. All experiments were run on 4-8 GPU nodes and took on the order of a few days depending on the particular experimental configuration.

\subsection{Evaluation metrics}
\myparagraph{FID justification for depth experiments.} For consistency with the other experiments, we continue to report FID metrics for the depth datasets as the protocol is an emerging standard comparing image distributions. The rationale is that while the pretrained InceptionV3 domain gap may shift the absolute ranges of scores, relative differences in scores should still give some insights into the performance of different models. Therefore, we supplement FID scores with standard depth-specific metrics for our experiments.

\subsection{Code}
Code is currently available at \url{https://github.com/alexandonian/contrastive-feature-loss}: please contact \texttt{andonian@mit.edu} for further details.
\section{Results using different encoder backbone architectures}

\begin{table*}[h!]
    \centering
    \small
    \begin{tabular}{lccccccc}
    \toprule
    \multicolumn{5}{c}{\textbf{Method}} & \multicolumn{3}{c}{\textbf{Performance}} \\
    \cmidrule(lr){1-5} \cmidrule(lr){6-8}
    Backbone & NCE variant    & GAN             & Pretrained $F$ & Frozen $F$ & FID $\downarrow$ & mAP $\uparrow$ & Acc $\uparrow$ \\
    \hline
    \multirow{4}{*}{\shortstack[l]{ImageNet-pretrained \\ ResNet50}}
                      & Standard       & \greencheck &  \greencheck    & \greencheck & 57.3      & \bf{64.3} & \bf{82.2} \\
                      & Standard       & \greencheck &  \redx          & \redx       & \bf{51.9} & 58.8      & 81.5      \\
                      & Bidirectional  & \greencheck &  \greencheck    & \greencheck & 63.3      & 62.0      & 82.1      \\
                      & Bidirectional  & \redx       &  \greencheck    & \greencheck & 80.3      & 60.3      & 81.6      \\
    \cdashline{1-8}
    \multirow{4}{*}{\shortstack[l]{6 Layer CUT \\ ConvEncoder}}
                      & Standard       & \greencheck &  \redx          & \redx       & 60.6      & 58.7      & 81.2      \\
                      & Standard       & \redx       &  \redx          & \redx       & 117.0     & 48.8      & 78.0      \\
                      & Bidirectional  & \greencheck &  \redx          & \redx       & 76.6      & 62.0      & 81.9      \\
                      & Bidirectional  & \redx       &  \redx          & \redx       & 86.4      & 61.5      & 81.9      \\
    \bottomrule
    \end{tabular}
    \caption{Performance on Cityscapes dataset using different backbone feature encoders. Surprisingly, the ImageNet-pretrained ResNet50 backbone is capable of achieving the best FID score of all the backbones, variants and baselines investigated in this work. The lower half of the table shows that feature encoders without pretraining can learn useful representations over the course of joint training with the generator.}
    \label{tab:additional_results}
\end{table*}

We wish to show that our method's effectiveness does not depend on a particular choice of backbone architecture or feature space. To this end, we evaluate our method's performance using two additional architectures: the popular ResNet50 architecture and a generic 6-layer CNN. For ResNet50, we extract features from the output of the first maxpool layer, and \texttt{layers1}$\dots$\texttt{layer4} as defined by the standard ResNet50 implementation in \texttt{torchvision.models}.

The second feature extraction network for our contrastive loss uses the same architecture as CUT \cite{park2020contrastive}, which consists of a simple 6-layer convolutional network followed by a linear projection and L2-normalization.
Inputs are first resized to 256x256 pixel resolution via bilinear interpolation.
Convolutional layers are followed by a spectral instance normalization layers.
Features are extracted from the first 5 layers as well as the raw RGB input pixels.
Here, we only adopt the specific architecture of the convolutional encoder network present in CUT~\cite{park2020contrastive}, while all other details pertaining to the NCE loss formulation remain the same as the main paper.

The results of these experiments and ablations shown in Table \ref{tab:additional_results} indicate that our method works effectively with additional backbone networks. In particular, an ImageNet-pretrained ResNet50 encoder can, in fact, be used to achieve an FID score (51.9) that \textbf{outperforms} both the best of our method using a VGG backbone and the SPADE~\cite{park2019SPADE} method. Similarly, the ResNet50 backbone also achieves segmentation metrics that rival the VGG based approach. 

Given the architecture of the 6 layer CUT encoder, pretrained ImageNet weights are not available and so these experiments are performed with an unfrozen encoder that is trained jointly with the generator. The lower half of Table \ref{tab:additional_results} shows that strong segmentation metrics (mAP and pixel Acc.) can be achieved \emph{even with a lightweight encoder without the need for any pretraining on large datasets}. Consistent with previous experiments, the small FID performance penalty incurred by forgoing expensive pretraining can be significantly reduced by employing the bidirectional NCE loss variant over the standard NCE loss and further reduced by including a GAN discriminator loss.

\section{Additional qualitative examples}

\begin{figure*}
    \centering
    \includegraphics[width=0.8\linewidth]{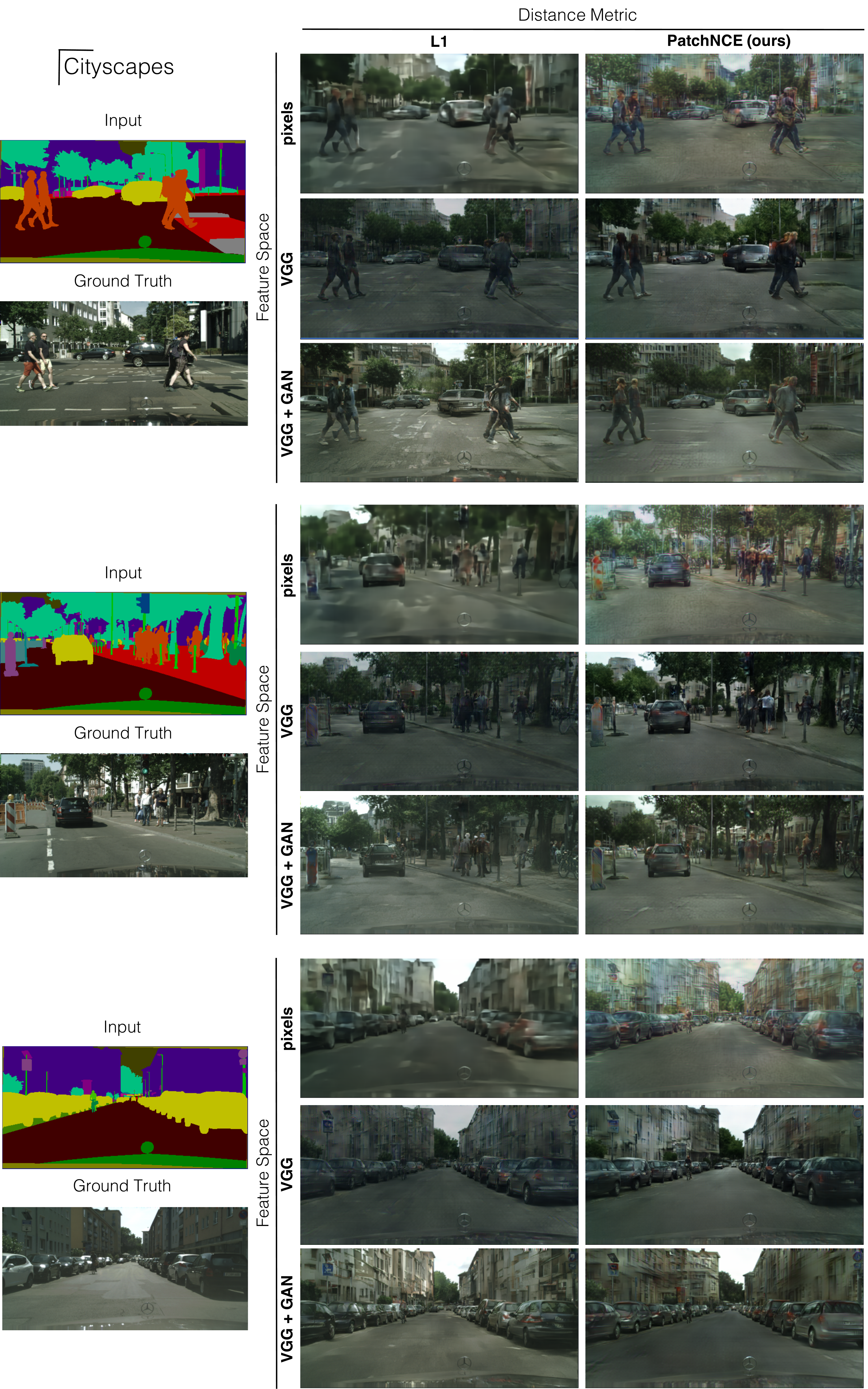}
    \caption{We show additional qualitative comparison results, extending the Cityscapes examples of Figure 3 and 4 of the main paper.}
    \label{fig:cityscapes_qual}
\end{figure*}

\begin{figure*}
    \centering
    \includegraphics[width=0.8\linewidth]{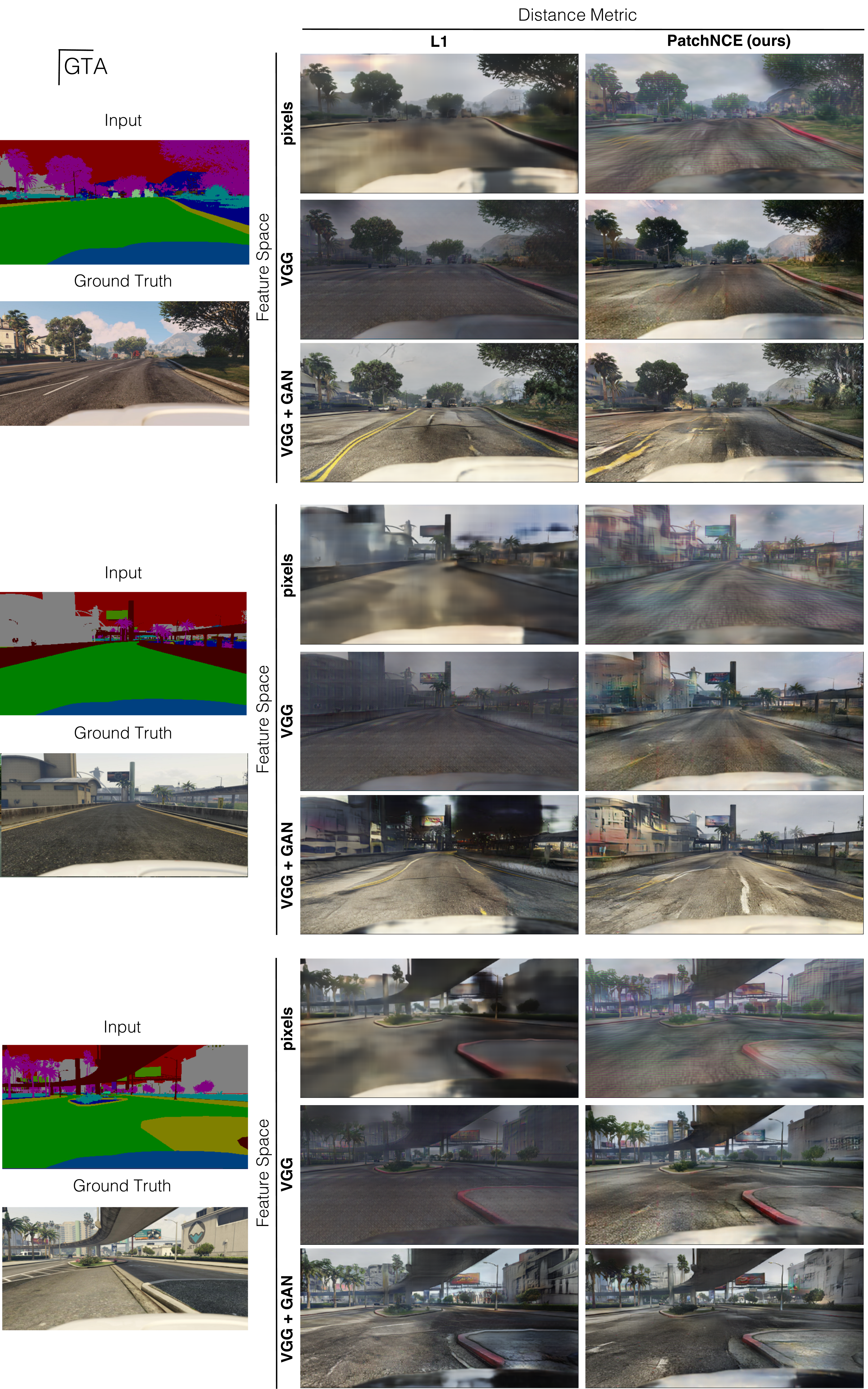}
    \caption{We show additional qualitative comparison results, extending the GTA examples of Figure 3 and 4 of the main paper.}
    \label{fig:gta_qual}
\end{figure*}

\begin{figure*}
    \centering
    \includegraphics[width=1.0\linewidth]{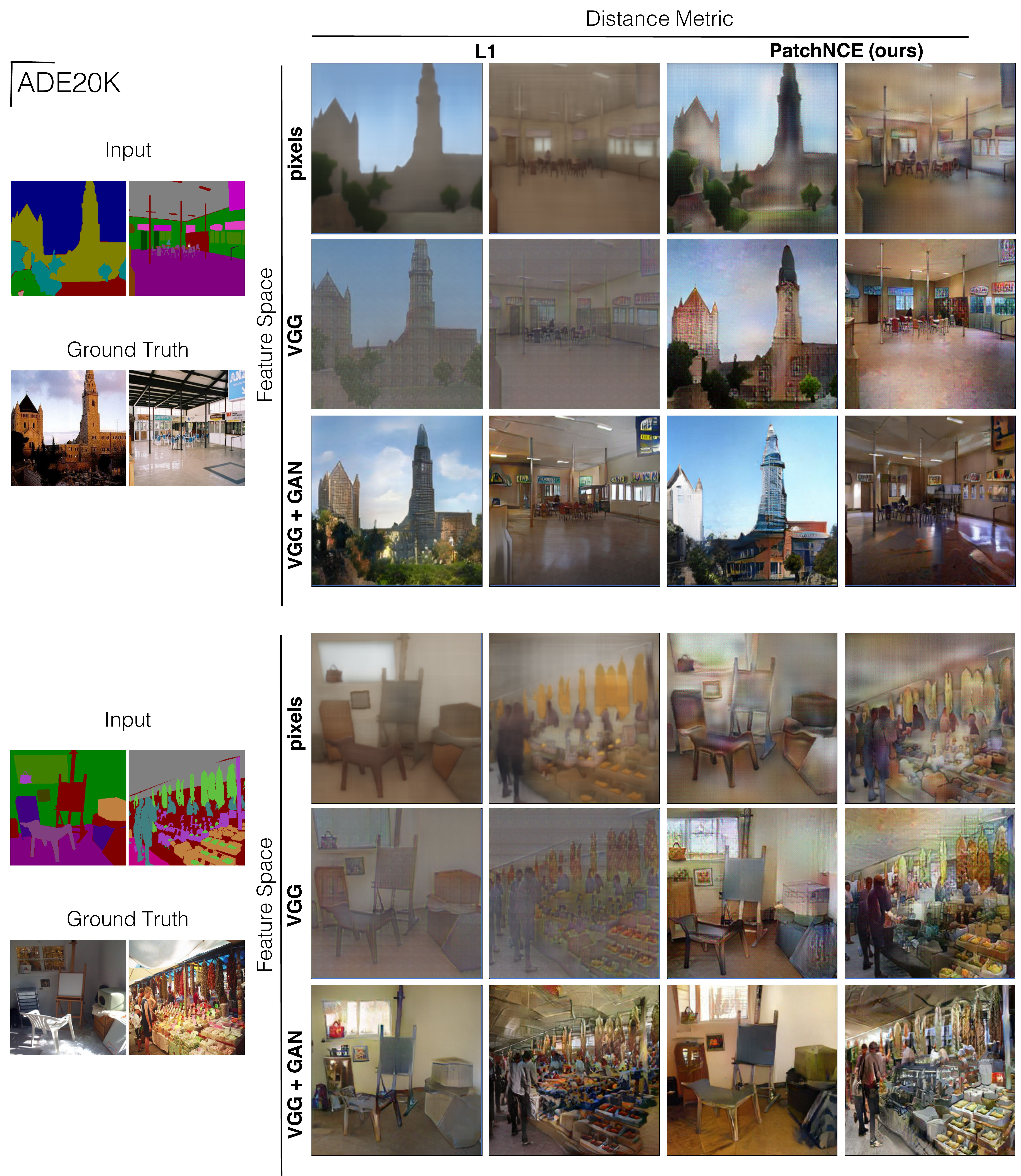}
    \caption{We show additional qualitative comparison results, extending the ADE20K examples of Figure 3 and 4 of the main paper.}
    \label{fig:ade20k_qual}
\end{figure*}

We show more qualitative comparisons to Figure 3 and 4 of the main paper, in Figure~\ref{fig:cityscapes_qual},~\ref{fig:gta_qual},~\ref{fig:ade20k_qual}, (this supplemental).

\section{Comparing the PatchNCE loss to GANs}
\begin{figure*}[h]
    \centering
    \includegraphics[width=0.9\linewidth]{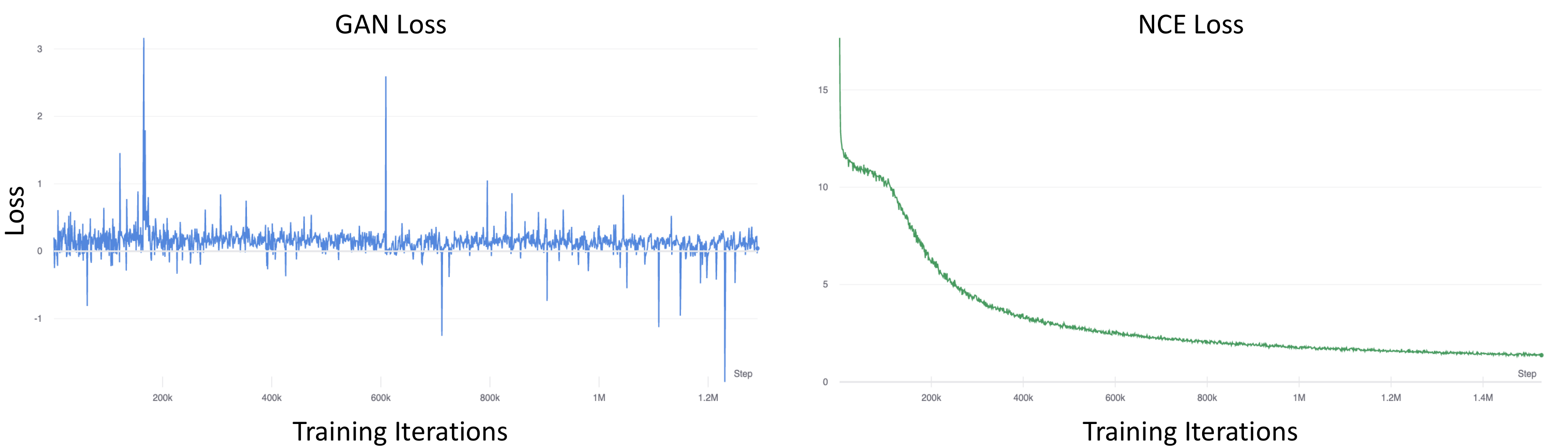}
    \caption{Representative loss trajectories over the course of training for GAN loss (left) and PatchNCE loss (right). GAN losses tend to be noisy with no clear indication of training progress. In contrast, the PatchNCE loss exhibits a clear downward trajectory under normal training circumstances, thus potentially serving as a strong diagnostic/debugging signal for practitioners.}
    \label{fig:loss_comparison}
\end{figure*}

Our PatchNCE loss aims to be a replacement for the L1 regression loss, while complementary to the GAN loss. Still, the output images produced by the PatchNCE loss bear similar characteristics to the GAN-generated images, such as color saturation and sharpness (Figure~\ref{fig:cityscapes_qual},~\ref{fig:gta_qual}, and~\ref{fig:ade20k_qual}). Also, in quantitative metrics (Table 1 of the main paper), using the PatchNCE loss alone without GAN often achieves higher segmentation scores than the GAN + L1 in the VGG feature space, at moderate sacrifice in FIDs. Therefore, the PatchNCE loss may be used without the GAN loss and still produce useful outputs for downstream tasks like domain adaptation. Here we discuss the benefits of using the PatchNCE loss without the GAN loss in image synthesis tasks.

\myparagraph{Training stability.}
Training GANs can be notoriously difficult.
Adversarial training is a highly dynamic characterized by instability, particularly in the discriminator network.
Models may never converge and mode collapses are common.
A growing body of research has focused of improvements to the GAN training procedure to impart stability, which include techniques such as ``progressive growing'' of the outputs, modifications to the GAN objective through gradient penalties/clipping, and particular normalization layers.
Still, debugging GANs can be difficult because objective is inherently adversarial and loss trajectories tend not to be very informative, as shown in figure \ref{fig:loss_comparison} (left). In contrast, our method trains the generator and feature loss in cooperative fashion, which is reflected in a smooth and interpretable loss trajectory (\ref{fig:loss_comparison}, right). As a consequence, we have empirically found that our method can sustain learning rates approximately 5 times larger for a given batch size thus providing an opportunity to increase convergence speed and reduce training time.

\myparagraph{Simpler implementation.}
In modern GAN implementations, training steps alternate between updating the generator and updating the discriminator, requiring two separate optimizers with potentially two different learning rates and additional choices such as updating one network more frequently than the other as with TTUR as seen in \cite{zhang2019self}.
Our method's training loop is conceptually simpler, and we find that both the generator and encoder can share the same optimizer and hyperparameters while making updates to both networks in a single gradient update. In this way, we eliminate the two step procedure for updating the generator and discriminator separately. Our loss can be incorporated into a full objective nearly as easily as it would be to incorporate an L1 loss term.

\myparagraph{Speed and memory comparisons.}
The results in section 2 demonstrate that our method is robust to choice of encoder. As such, in addition to removing the discriminator network, one can choose a lighter, more modern and efficient networks to serve as the contrastive encoder in our formulation to accommodate the compute and memory constraints of a given circumstance. For example, replacing the pretrained VGG network with ResNet50 reduced the number of weight parameters by over 6x (143.6M vs 23M) with very marginal performance penalties. Further memory reductions are achieved with the 6 layer CUT encoder (9.1M parameters) for a total of nearly 16x few parameters, matching segmentation metric scores and only mild FID score degradation. 

\myparagraph{Performance comparisons.}
Our method provides clear advantages in achieving better segmentation metrics such as pixel accuracy and mAP, while still maintaining competitive FID scores.

\end{document}